\documentclass[10pt,twocolumn,letterpaper]{article}

\usepackage{iccv}              

\usepackage[accsupp]{axessibility}  

\usepackage{xspace}
\definecolor{iccvblue}{rgb}{0.21,0.49,0.74}
\usepackage[pagebackref,breaklinks,colorlinks,allcolors=iccvblue]{hyperref}

\def\paperID{1349} 
\def\confName{ICCV}
\def\confYear{2025}

\title{\model: 3D Generative Avatar Prior for Monocular Gaussian Avatar Reconstruction}

\author{\vspace{.05cm} Zijian Dong$^{1,4*}$ \quad Longteng Duan$^{1*}$ \quad Jie Song$^{3}$ \quad Michael J. Black$^{4}$  \quad Andreas Geiger$^{2}$ \\
\vspace{.05cm}$^1$ETH Z{\"u}rich \quad $^2$University of Tübingen, Tübingen AI Center \\ $^3$HKUST(GZ)$\&$HKUST \quad $^4$Max Planck Institute for Intelligent Systems, T{\"u}bingen
}

\newcommand{\figref}[1]{Fig.~\ref{#1}}
\newcommand{\secref}[1]{Section~\ref{#1}}

\newcommand{\tabref}[1]{Table~\ref{#1}}

\makeatletter
\DeclareRobustCommand\onedot{\futurelet\@let@token\@onedot}
\def\@onedot{\ifx\@let@token.\else.\null\fi\xspace}

\makeatother

\newcommand{\boldparagraph}[1]{\vspace{0.2cm}\noindent{\bf #1:}}

\newif\ifshowcomments
\showcommentstrue 

\ifshowcomments
    \newcommand{\oh}[1]{ \noindent {\color{red} {\bf OH:} {#1}} }
    \newcommand{\ag}[1]{ \noindent {\color{red} {\bf AG:} {#1}} }
    \newcommand{\mjb}[1]{ \noindent {\color{red} {\bf MJB:} {#1}} }
    
    \newcommand{\xc}[1]{ \noindent {\color{red} {\bf XC:} {#1}} }
    \newcommand{\zj}[1]{ \noindent {\color{red} {\bf Zijian:} {#1}} }
    \newcommand{\lt}[1]{ \noindent {\color{red} {\bf Longteng:} {#1}} }

 \else
    \newcommand{\oh}[1]{\unskip}
    \newcommand{\ag}[1]{\unskip}
    \newcommand{\mjb}[1]{\unskip}
    \newcommand{\jy}[1]{\unskip}
    \newcommand{\xc}[1]{\unskip}
    \newcommand{\zj}[1]{\unskip}
    \newcommand{\lt}[1]{\unskip}

\fi   

\newcommand{\model}{MoGA\xspace}

\begin{document}

\twocolumn[{%
\renewcommand\twocolumn[1][]{#1}%
\maketitle
\vspace{-3.5em}

\begin{center}
    \includegraphics[width=\textwidth,trim=0 0 0 0, clip]{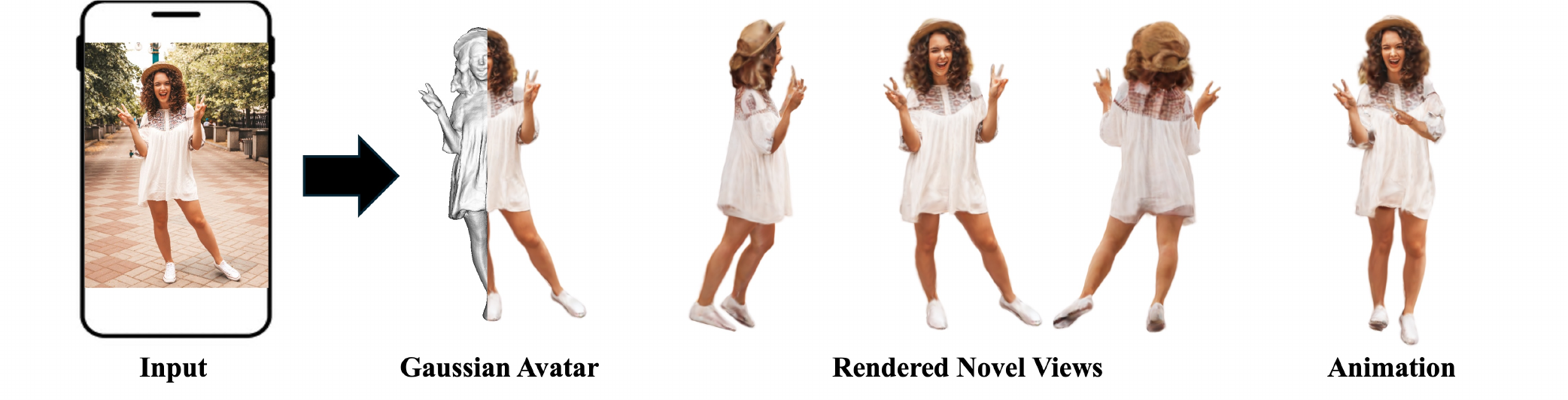}
    \vspace{-2.5em}
    \captionof{figure}{
    We propose \model, a method to genereate high-fidelity Gaussian avatars from a single image. Left: A challenging in-the-wild example. Middle: Unlike previous methods that struggle with such cases (\figref{fig:comparison_wild}), \model enables 3D-consistent full-body novel view synthesis and detailed geometry extraction. Right: Our reconstructed Gaussian avatar supports animation without any post-processing.}  
    \label{fig:teaser}
\end{center}

}
]


\def\thefootnote{*}\footnotetext{Equal contribution}
\begin{abstract}

\vspace{-1em}
We present \model, a novel method to reconstruct high-fidelity 3D Gaussian avatars from a single-view image. 
The main challenge lies in inferring unseen appearance and geometric details while ensuring 3D consistency and realism. 
Most previous methods rely on 2D diffusion models to synthesize unseen views; however, these generated views are sparse and inconsistent, resulting in unrealistic 3D artifacts and blurred appearance.
To address these limitations, we leverage a generative avatar model, that can generate diverse 3D avatars by sampling deformed Gaussians from a learned prior distribution.
Due to limited 3D training data, such a 3D model alone cannot capture all image details of unseen identities.
Consequently, we integrate it as a prior, ensuring 3D consistency by projecting input images into its latent space and enforcing additional 3D appearance and geometric constraints.
Our novel approach formulates Gaussian avatar creation as model inversion by fitting the generative avatar to synthetic views from 2D diffusion models.
The generative avatar provides an initialization for model fitting, enforces 3D regularization, and helps in refining pose.
Experiments show that our method surpasses state-of-the-art techniques and generalizes well to real-world scenarios. Our Gaussian avatars are also inherently animatable. 
For code, see \url{https://zj-dong.github.io/MoGA/}

\end{abstract}    


\section{Introduction}
\label{sec:intro}


Animatable and realistic avatar creation enables many applications in AR/VR, movies, and the gaming industry. It is hard to scale up this process since previous traditional approaches~\cite{joo2015panoptic} require expensive multi-view systems and highly-specialized expertise to craft the avatar. To make digital avatars widely available and consumer-friendly, it is essential to develop methods for creating an avatar from in-the-wild images. However, this problem is very challenging due to the ill-posed nature of the monocular setting, which causes ambiguities in the appearance, depth, and body poses.  

Embracing the challenging problem,  previous methods~\cite{xiu2022icon, saito2019pifu, saito2020pifuhd}  learn a large network to predict explicit or implicit 3D representations from 2D pixel-aligned features. These methods are trained on small-scale datasets due to the limited amount of available 3D data; this greatly restricts their generalization to diverse human poses and clothing styles. More recently, to enable more powerful generalization ability,
some methods~\cite{li2024pshuman, xue2024human, ho2024sith, zhang2024sifu} leverage a multi-view diffusion model to hallucinate back and side views. 
Despite impressive performance, most existing multi-view diffusion models can only generate very sparse views with high resolution due to memory constraints during training~\cite{kant2025pippo}.  The sparsity of generated views leads to artifacts in self-occluded regions or unobserved side views. Furthermore, since such methods rely heavily on 2D priors from 2D diffusion, the generated multi-view imagery often lacks 3D consistency~\cite{qian2023magic123, liu2023one}, which results in blurry appearance in 3D.  
To prevent the reconstruction of unnatural bodies, some methods~\cite{li2024pshuman, huang2024tech, chen2024generalizable,zhang2024sifu, ho2024sith} leverage parametric body models like SMPL~\cite{loper2023smpl} as full-body priors. Although this helps avoid abnormal shapes, it is restricted by the fixed topology and minimally-clothed SMPL body shape and cannot provide a 3D appearance prior. 

To address these limitations, we propose a novel method, \model (Moncular Gaussian Avatar), that reconstructs a high-fidelity 3D Gaussian avatar from a single-view image (Fig.~\ref{fig:teaser}). At its core, our approach leverages a generative 3D avatar model as a powerful human body prior. Unlike the SMPL body prior, our model captures not only detailed geometry but also realistic human appearance, including hair and clothing, using deformed Gaussians. We harness multi-view diffusion to infer unseen views while ensuring 3D consistency and realism. This is achieved by projecting the synthetic images back to the learned latent space of the generative avatar and applying additional constraints.
The creation of the Gaussian avatar is then formulated by fitting the generative avatar to these generated views. During this process, our generative avatar model plays a pivotal role in three ways: (i) \textbf{Initialization:} Sampling from the learned avatar prior enables a meaningful initialization
of geometry and appearance for fitting, helping avoid local minima in few-shot reconstruction.  (ii) \textbf{Regularization:} Rather than relying solely on inconsistent synthetic images, the avatar prior enforces strong 3D constraints, ensuring view consistency and preventing unrealistic artifacts. (iii) \textbf{Pose Optimization:} This avatar prior enables us to refine human and camera poses through an effective photometric rendering loss, improving alignment accuracy and reconstruction fidelity over baselines.

More specifically, we represent human bodies and clothing in canonical space using 2D Gaussian Splatting~\cite{huang20242dgs}, anchored to a parametric body template~\cite{pavlakos2019smplx}, and integrate it with an efficient deformation module~\cite{dong2023ag3d}. 
To enable generation, we model every human subject via a per-subject latent code and a shared decoder to interpret the latent code into Gaussian features. 
%
%
This generative avatar prior is learned through a single-stage pipeline~\cite{chen2023single} that jointly optimizes the latent code, the shared decoder, and a latent diffusion model in the latent space. At test time, we first employ image-guided sampling~\cite{chen2023single} to obtain a meaningful latent code. Given this initialization, we perform model inversion to compute the latent code for a novel target identity while freezing the decoder and learned diffusion model. During this process, the diffusion model serves as a constraint in the latent space via a score distillation sampling loss~\cite{poole2022dreamfusion}. Throughout the fitting procedure, both the avatar model and camera/human pose parameters are optimized in an alternating manner to correct abnormal poses. 

We experimentally demonstrate that our method significantly outperforms previous state-of-the-art methods both quantitatively and qualitatively (\tabref{tab:comparison} and \figref{fig:comparison}). The resulting Gaussian avatar has better 3D consistency and realism (\figref{fig:comparison}), and adapts better to model complex structures like hair (\figref{fig:topology}). In summary, we contribute:

\begin{itemize}

 \item An optimization-based model fitting framework to reconstruct a Gaussian avatar from a single image, by fitting a 3D generative avatar to synthetic images generated by 2D multi-view diffusion. 
 
 \item A generative 3D Gaussian avatar prior that enables reconstructing a deformed Gaussian avatar from sparse and inconsistent generated images, by providing crucial support for meaningful initialization, regularization, and pose refinement during model fitting. 

 \item Generalization to in-the-wild outdoor images with challenging pose and clothing. The resulting avatar can be animated without post-processing. 
\end{itemize}
Code and models are available at \url{https://zj-dong.github.io/MoGA/}.



\section{Related Work}
\label{sec:related_work}

\subsection{Single-view Human Reconstruction}
Creating realistic avatars from a {\em single RGB image} is a challenging problem.  PIFu~\cite{saito2019pifu} pioneered a data-driven pipeline to learn a mapping from 2D pixel-aligned features to 3D implicit functions. 
More recent work builds upon this idea and improves the geometry by leveraging normal guidance and parametric human body models  \cite{xiu2022icon, saito2020pifuhd, alldieck2022photorealistic, liu2023one, liu2023zero}. However, given only a frontal view, these methods struggle to reconstruct realistic full-body texture because of the unobserved back side.  More recently, methods use 2D diffusion models to hallucinate the back view and incorporate this into the reconstruction pipeline \cite{ho2024sith, zhang2024sifu}. 
However, with only two observations, these methods suffer from strong artifacts in side-views. 
Several methods go further by leveraging a modified multi-view diffusion process to generate more views, improving rendering quality and geometry \cite{xue2024human, li2024pshuman, zhang2024sifu}. 
Unfortunately, 2D diffusion often produces synthetic images that are inconsistent in 3D.
To address this, PSHuman \cite{li2024pshuman} and SIFU \cite{zhang2024sifu} leverage SMPL \cite{loper2023smpl, pavlakos2019smplx} as a 3D template to regularize the reconstruction. However, since SMPL does not provide an appearance prior and only has a minimally clothed body shape, the methods struggle when the clothing and hair differ from the SMPL body topology.
Human3Diffusion~\cite{xue2024human} uses a  3D diffusion model to guide sampling of a multi-view diffusion model, but is limited by the low resolution of the diffusion model. 
In contrast to the prior work, we combine an expressive generative 3D avatar model with synthetic 2D images generated from a multi-view diffusion model, achieving better reconstruction quality and robustness to self-occlusion.




\subsection{3D Avatar Generation}
Several methods~\cite{dong2023ag3d, hong2022eva3d, noguchi2022enarfgan, abdal2024gsm} leverage 3D-aware GANs \cite{chan2021pi, chan2022eg3d, schwarz2020graf} to generate 3D humans from 2D image collections. 
The main idea is to leverage 3D human models~\cite{loper2023smpl, pavlakos2019smplx} to learn a 3D human GAN with an adversarial loss. Several techniques are then developed to improve geometric quality~\cite{dong2023ag3d}, the face region~\cite{dong2023ag3d, hong2022eva3d}, deformation~\cite{dong2023ag3d} and efficiency~\cite{abdal2024gsm}. 
Despite impressive results, these methods are all trained with 2D image discriminators that are unable to reason about cross-view relationships~\cite{chen2023single}, making it challenging to exploit multi-view data. 
Recent 3D diffusion models \cite{ho2020denoising, song2020score,chen2023single,shue20233d} show better generation capabilities due to their more expressive and high-dimensional latent space. 
Leveraging this, recent work \cite{hu2024structldm, chen2023primdiffusion, e3gen} parameterizes the human via primitives~\cite{chen2023primdiffusion} or a structured latent code~\cite{hu2024structldm} and learns a diffusion model in the latent space for unconditional generation. By leveraging multi-view data, these methods acheive 3D consistency in appearance generation.  
Unlike these methods we leverage a 3D generative avatar model for single-view avatar reconstruction.  
Although reconstruction from single images is possible through GAN inversion~\cite{chan2022eg3d}, the quality is limited by the expressiveness of the latent space, making it hard to apply to in-the-wild images. Rodin~\cite{wang2023rodin} and its extensions~\cite{zhang2024rodinhd} employ an image-conditioned diffusion model for few-shot face reconstruction. 
These are trained on synthetic images and struggle to generalize to the real-world.  
We address the more challenging problem of reconstructing full-body avatars with diverse poses and clothing styles from in-the-wild images.

\subsection{Gaussian Avatar}
Representations for 3D avatars include 3D meshes~\cite{loper2023smpl, pavlakos2019smplx, alldieck2018video}, implicit functions \cite{peng2021neural, dong2022pina, lu2024avatarpose, guo2023vid2avatar, jiang2022instantavatar, weng2022humannerf}, and point clouds \cite{zheng2023pointavatar, tang2023human}. 
Recently, 3D Gaussian Splatting \cite{kerbl20233dgs, huang20242dgs} has gained attention due to its high rendering quality and efficiency. 
Many methods~\cite{hu2024gaussianavatar, qian20243dgs, zhao2024chase, yuan2024gavatar, moon2024expressive} learn 3D Gaussian avatars from monocular videos. 
Despite strong performance, these methods typically fail when the number of observations becomes sparse. GPS-Gaussian~\cite{zheng2024gps} proposes a generalizable multi-view human Gaussian model with high-quality rendering, but it needs relatively dense views (16) and accurate camera poses. 
Concurrent methods \cite{pan2025humansplat, chen2024generalizable} leverage a pre-trained transformer to predict the 3D Gaussians from a single-view image, but struggle to reconstruct details and tend to generate artifacts on faces. In contrast, our method achieves a more detailed and realistic Gaussian avatar from only a single-view image. A detailed comparison with \cite{pan2025humansplat, chen2024generalizable} is not possible since the models or code have not been released at the time of writing.








\section{Method}
\begin{figure*}[t]
    \centering
    \includegraphics[width=\textwidth]{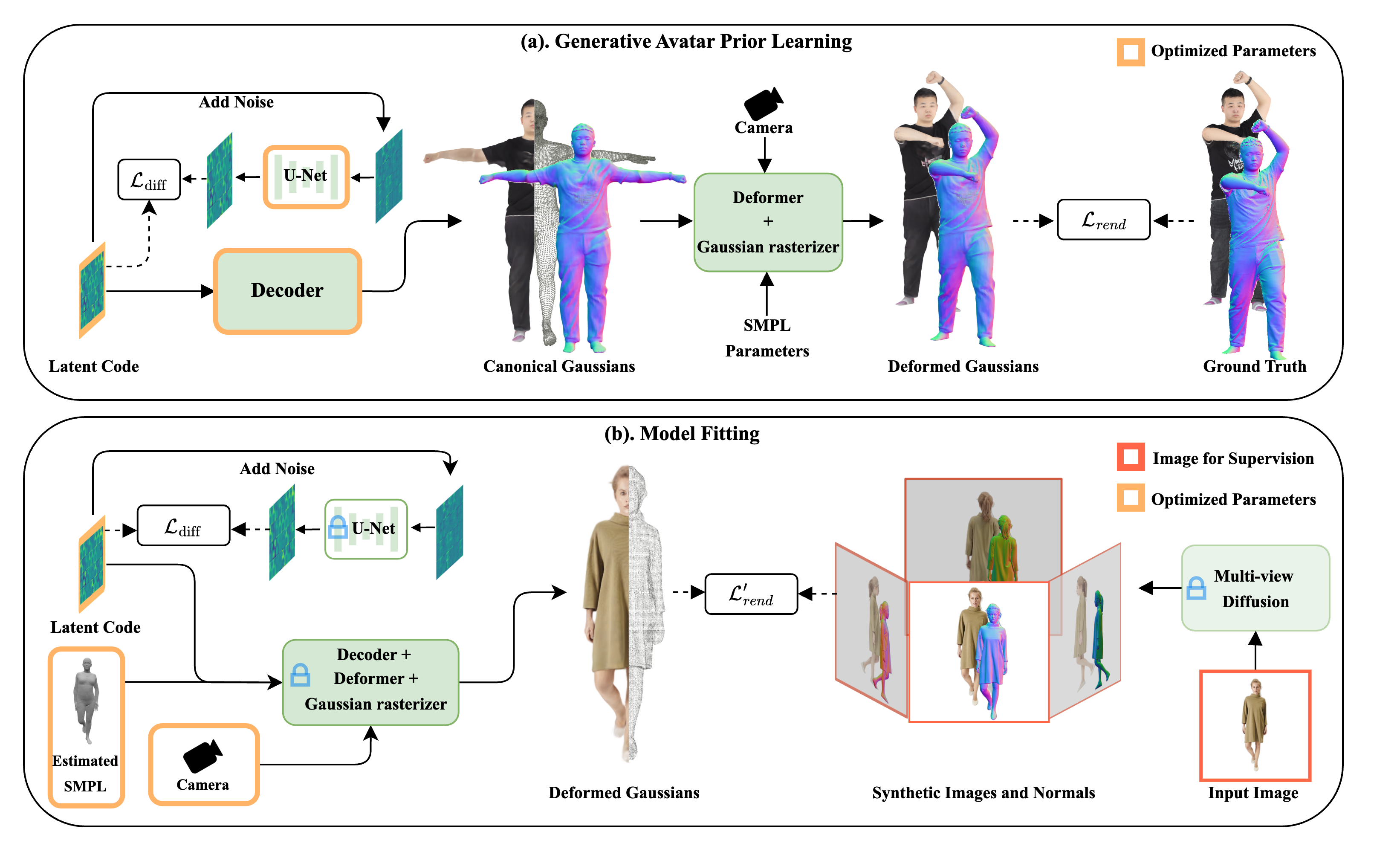}
    \caption{\textbf{Method Overview.} \emph{Generative Avatar Prior Learning:}
Our 3D human generator creates the appearance and geometry in canonical space represented by 3D Gaussians and leverages an efficient deformation module to deform these into posed space for Gaussian rasterization. To learn this generative avatar model from a 3D human dataset, we utilize a single-stage training pipeline that jointly optimizes a Gaussian auto-decoder (including a per-subject latent code and a shared decoder) and a latent diffusion model. \emph{Model Fitting:} At test time, we fit the learned generative avatar to synthetic images generated from a pretrained multi-view diffusion model. During this process, we first initialize the latent code by image-guided sampling and perform model inversion to compute the latent code while freezing the decoder and learned diffusion model. Both the avatar model and camera/human pose parameters are optimized in an alternating manner to correct abnormal poses. }
    \label{fig:method}
\end{figure*}



Given a single-view image, our goal is to reconstruct a high fidelity 3D Gaussian avatar. To address this ill-posed problem, our key idea is to leverage a generative 3D avatar model as a human prior and fit this generative model to synthetic images generated by multi-view diffusion. An overview of our method is shown in \figref{fig:method}.

We first introduce an efficient and articulation-aware 3D human generator (\secref{sec:avatar_training} and \figref{fig:method}(a)), which generates the appearance and shape in canonical space and leverages a deformation module to deform these into posed space. To learn this generator, we utilize a single-stage training pipeline~\cite{chen2023single} that jointly optimizes a Gaussian auto-decoder and a latent diffusion model. 

At test time, we fit the learned generative avatar model to 6 synthetic images generated using a pre-trained multi-view diffusion model (\secref{sec:fitting} and \figref{fig:method}(b)).  We show that the learned generative prior improves the performance by providing a good initialization and regularization for fitting to handle the inconsistency between synthetic views. During the fitting process, we optimize both the avatar model and the camera/human pose parameters alternately to correct abnormal poses.

\subsection{Generative Avatar Prior Training}
\label{sec:avatar_training}

\subsubsection{Canonical Gaussian Representation}

To achieve high quality reconstruction of both appearance and geometry, we employ 2D Gaussian splatting~\cite{huang20242dgs} to represent the appearance and geometry of the avatar generated in the canonical space. Motivated by GGHead~\cite{kirschstein2024gghead} and Relightable Gaussian Codec Avatars~\cite{saito2024relightable}, we parameterize 2D Gaussians $\mathcal{G}$ on a UV map $U$ of a template mesh~\cite{pavlakos2019smplx}.  Here, each Gaussian primitive $\mathcal{G}_k$ is parameterized by five attributes: an opacity $\sigma_k \in \mathbb{R}$, a Gaussian center $\mu_k \in \mathbb{R}^3$, RGB color $c_k \in \mathbb{R}^3$ for simplicity, a scale vector $s_k \in \mathbb{R}^2$ for 2D Gaussian Splatting, and a rotation matrix $R_k$ represented by the axis angle vector $r_k \in \mathbb{R}^3$. Finally, based on the UV mapping, we can represent the Gaussian attributes $\mathcal{G}$ with a 2D UV map $U \in \mathbb{R}^{256 \times 256 \times 12}$. 

To better leverage the template body prior, we model the Gaussian center $\mu_k$, the scale vector $s_k$, and the rotation $r_k$ as a residual from the canonical SMPL-X body \cite{pavlakos2019smplx}:
\begin{equation}
\begin{aligned}
    \mu_k &= \hat{\mu}_k + \delta_{\mu k} \\
    s_k &= \hat{s}_k \cdot \delta_{s k} \\
    r_k &= \hat{r}_k \cdot \delta_{r k} .
\end{aligned}
\end{equation}
Here, $\hat{\mu}_k$, $\hat{s}_k$, and $\hat{r}_k$ are the initial center, scale, and rotation of the Gaussian primitives, which are obtained from the SMPL-X mesh; for more details, see Sup.~Mat. 
Finally, we predict the offset value $\delta_{\mu k}$, $\delta_{s k}$, $\delta_{r k}$ to represent Gaussian attributes. 

To make our model generalize to various people, we learn a shared auto-decoder across all the training people. For each identity, we model each person by a small compressed latent code $X_i \in \mathbb{R}^{64 \times 64 \times 32}$ and then decode the latent code to final UV map $U_i$ using the shared CNN decoder. More details of the decoder can be found in Sup.~Mat.

\subsubsection{Deformer}
To enable animation and learn from posed images, we use a deformer to transform the avatar $\mathcal{G}$ from the canonical space into posed space. For each Gaussian primitive $\mathcal{G}_k$, the deformed Gaussian center and rotation matrix $\mu'_k$ and $R'_k$ are computed as:
\begin{equation}
    \mu'_k = T\mu_k,  R'_k = TR_k, \text{ where } T=\sum_{i=1}^{n_b} w_i B_i .
\end{equation}
Here $n_b$ is the number of joints, $B_i$ is the bone transformation matrix for joint $i \in \left\{1, ..., n_b \right\}$, and $w_i$ is the skinning weight, which determines the influence of the motion of each joint on $\mu_k$. Following AG3D~\cite{dong2023ag3d}, the skinning weight is represented as a low-resolution voxel grid. More details can be found in Sup.~Mat.




\subsubsection{Rendering}
After we obtain the deformed Gaussian attributes, we perform 2D Gaussian splatting as in \cite{huang20242dgs}. For each pixel $\textbf{x}=(x,y)$, the pixel color is obtained by:
\begin{equation}
    c(\textbf{x}) = \sum_{i=1}^N c_i \mathcal{G}_i(\textbf{x}) \sigma_i \prod_{j=1}^{i-1}(1 - \sigma_j \mathcal{G}_j(\textbf{x}))
\end{equation}
where $c_i$ is the color of the $i$-th projected 2D Gaussian primitive sorted by depth. To render normal maps, we replace the color $c_i$ with the normal of the Gaussian primitives. $\sigma_i$ represents the opacity values. $\mathcal{G}(\textbf{x})$ is the evaluated 2D Gaussian value. More details of the evaluation of $\mathcal{G}(\textbf{x})$ can be seen in \cite{huang20242dgs}.

\subsubsection{Generative Avatar Training}

\figref{fig:method}(a) illustrates the training process of this generative avatar model. We leverage the latent diffusion model (LDM)~\cite{rombach2022high} to learn the generative prior in the latent space. Following SSDNeRF~\cite{chen2023single}, we adopt a single-stage training pipeline to jointly optimize our auto-decoder and LDM. The training objective is:
\begin{equation}
\label{eq:objective}
    \mathcal{L} = \lambda_{\text{rend}} \mathcal{L}_{\text{rend}}(\{X_i\}, \psi) + 
                    \lambda_{\text{diff}} \mathcal{L}_{\text{diff}}(\{X_i\}, \phi).
\end{equation}
Here $X_i$ is the latent feature code and $\psi$ and $\phi$ denote the parameters of the decoder and denoising U-Net respectively. $\lambda_*$ are the loss weights. $\mathcal{L}_{\text{rend}}$ and $\mathcal{L}_{\text{diff}}$ are the training objectives for the rendering and diffusion process. Compared to two-stage training~\cite{shue20233d, wang2023rodin}, the resulting learned latent space is smoother due to end-to-end optimization of the diffusion and decoder weights. The rendering loss is:
\begin{equation}
    \mathcal{L}_{\text{rend}}(\{X_i\}, \psi) = \lambda_{\text{l2}} \mathcal{L}_{\text{l2}} +                 \lambda_{\text{vgg}} \mathcal{L}_{\text{vgg}}+
                                        \lambda_{\text{reg}} \mathcal{L}_{\text{reg}} .
\end{equation}
$\mathcal{L}_{\text{l2}}$ is the L2 reconstruction loss between rendering and observations. 
Unlike \cite{chen2023single}, we compute the reconstruction loss on both RGB and normal images to improve the geometry quality. $\mathcal{L}_{\text{vgg}}$ is a perceptual loss based on the difference between the feature maps obtained from \cite{simonyan2014vgg} and the rendered image. We define $\mathcal{L}_{\text{reg}}=\| \delta_{\mu k} \|$  to prevent the predicted offset from being too large.

To train the diffusion model, similar to ~\cite{chen2023single}, we compute $\mathcal{L}_{\text{diff}}$
as:
\begin{equation}
\label{eq:diffusion}
    \mathcal{L}_{\text{diff}}\left( \{X_i\}, \phi \right) = 
            \mathop{\mathbb{E}}_{i,t,\epsilon} \left [
                        \frac{1}{2} w^{(t)} 
                        \left \| 
                            \hat{X}_i - X_i
                        \right\|^2
                        \right]
\end{equation}
where $\hat{X}_i$ is the denoised latent code with time step  
$t \sim \mathcal{U}(0, T)$, $w^{(t)}$ is an empirical time-dependent weighting function, and $\epsilon$ is the added noise. More details can be found in Sup.~Mat.

\subsection{Model Fitting}
\label{sec:fitting}

Equipped with the learned generative avatar prior,  we reconstruct a personalized avatar by fitting the generative model to synthetic views generated from multi-view diffusion.

\subsubsection{Multi-view Hallucination and Pose Estimation}

Since a single image is not enough for Gaussian reconstruction, we leverage a pre-trained multi-view diffusion model~\cite{li2024pshuman} to hallucinate 6 synthetic human images from a single image. After obtaining synthetic views, similar to \cite{shen2023x}, we leverage a human pose estimator to obtain the initial SMPL-X parameters. More details of preprocessing can be found in Sup.~Mat. 

\subsubsection{Rendering Objective}

To optimize the latent feature map, we define the rendering loss $ \mathcal{L'}_{\text{rend}}$ during inference as:
\begin{equation}
\begin{aligned}
    \mathcal{L'}_{\text{rend}}(\{X_i\}, \psi) = &\lambda_{\text{l2}} \mathcal{L}_{\text{l2}} + \lambda_{\text{vgg}} \mathcal{L}_{\text{vgg}} +
                                        \lambda_{\text{reg}} \mathcal{L}_{\text{reg}}  \\
                                       & + \lambda_{\text{nc}} \mathcal{L}_{\text{nc}} +
                                        \lambda_{\text{d}} \mathcal{L}_{\text{d}} .
\end{aligned}
\end{equation}
Here we use the same equation as in Eq.~\eqref{eq:objective} to calculate the L2 reconstruction loss $\mathcal{L}_{\text{l2}}$ and perceptual loss $\mathcal{L}_{\text{vgg}}$ between predicted images (normals) and generated synthetic images (normals). To improve the geometry of the avatar, we additionally add the normal consistency loss $\mathcal{L}_{\text{nc}}$ and the depth distortion loss $\mathcal{L}_{\text{d}}$ from ~\cite{huang20242dgs}. More details can be found in the Sup.~Mat.

\subsubsection{Prior-guided Optimization}
Since the multi-view diffusion model generates sparse and inconsistent images, solely relying on the rendering loss makes the result blurry and unrealistic.  Here, we tackle the problem by leveraging our pretrained generative avatar prior as a powerful human prior of 3D appearance and geometry. More concretely, this prior mainly contributes to three aspects including:

\paragraph{Initialization.} 
Random initialization of the latent code can sometimes cause the optimization to converge to a bad local minimum.
To solve this problem, we leverage our learned generative model to provide a good starting point for model fitting. To make our model generalizable to unseen test images, we follow \cite{chen2023single} to use image-guided sampling. More specifically, for a noisy code $X^{(t)}$ at every denoising step $t$, we additionally compute an approximated rendering gradient $g$ based on testing rendering loss $\mathcal{L'}_{\text{rend}}(X^{(t)})$ and add it to the denoised output $\hat{X}^{(t)}$ as an image-guided correction. 
More details of this computation can be found in Sup.~Mat. 




\paragraph{Regularization.}
Image-guided sampling provides a good initialization, but still cannot reconstruct all the details of test images. To solve this, we refine the sampled latent code by solving :
\begin{equation}
\label{eq:objective_test}
    \min_X \lambda_{\text{rend}} \mathcal{L'}_{\text{rend}}(X) + 
                    \lambda_{\text{diff}}' \mathcal{L}_{\text{diff}}(X).
\end{equation}
Here we optimize the diffusion loss defined in Eq.~\eqref{eq:diffusion} jointly with the rendering loss, while freezing the weights of the diffusion and decoder models. 
Unlike \cite{poole2022dreamfusion}, the diffusion model trained in UV feature space serves as a prior to regularize and inpaint the noisy and incomplete latent code during optimization. 

\begin{figure*}[h!]
    \centerline{    \includegraphics[width=\textwidth]{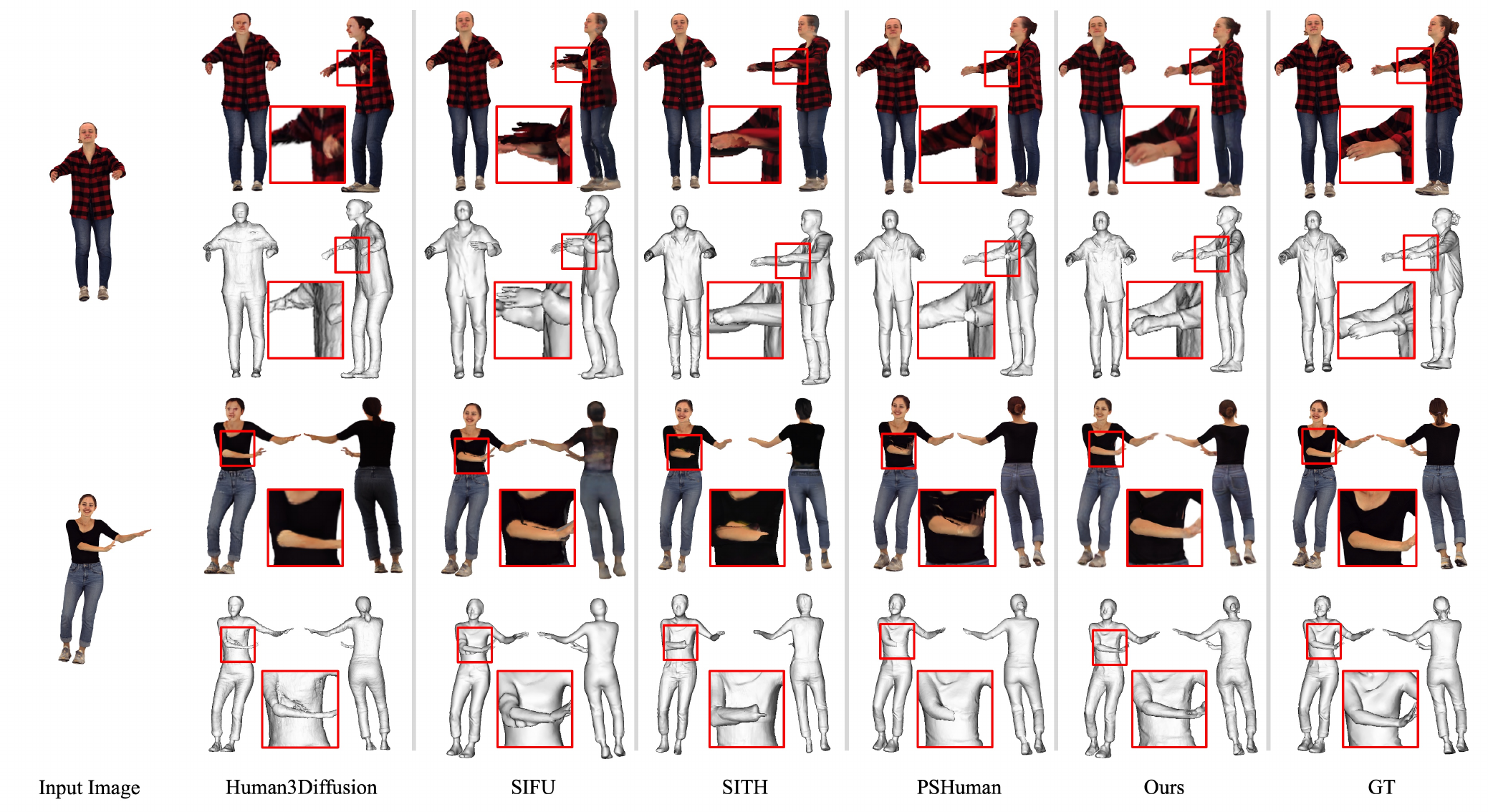}}
    \vspace{-0.1in}
    \caption{\textbf{Qualitative comparison to SotA methods on CustomHuman.} Our method achieves better image and shape quality,  enables 3D consistency in side views, and avoids unrealistic reconstruction due to self-occlusion.  }
    \label{fig:comparison}
\end{figure*}

\paragraph{Pose Optimization.}
The generated synthetic images from multi-view diffusion models are inconsistent, and this results in inaccurate camera and body pose estimation. Instead of using noisy 2D joint estimates~\cite{Bogo:ECCV:2016, pavlakos2019expressive}, we optimize pose parameters via a more effective photometric loss: 
\begin{equation}
    \mathcal{L}_{\text{pose}}  = \lambda_{\text{l2}} \mathcal{L}_{\text{l2}} +
                                \lambda_{\text{vgg}} \mathcal{L}_{\text{vgg}} +
                                \lambda_{\text{mask}} \mathcal{L}_{\text{mask}}
\end{equation}
where we combine a mask loss $\mathcal{L}_{\text{mask}}$ with predefined L2 loss $\mathcal{L}_{\text{l2}}$ and a perceptual loss. Here we optimize both SMPL parameters and camera poses by back-propagation. We optimize the latent code, camera and human poses in an alternating manner to avoid falling into locally suboptimal results. More details can be found in Sup.~Mat.

\section{Experiments}
In our experiments, we first compare our method to state-of-the-art (SotA) baselines on two public datasets and then test the generalization capability of our method on in-the-wild images. In addition, we provide an ablation study to investigate the importance of each component in our model.

\noindent\textbf{Datasets.}
We evaluate our proposed method on THuman2 \cite{tao2021function4d} and CustomHumans \cite{ho2023customhuman}. 
To test the generalization ability, we also collect some in-the-wild images from the Internet for qualitative comparison.  For details see Sup.~Mat.

\noindent\textbf{Metrics.}
Following previous methods~\cite{li2024pshuman, ho2024sith}, we evaluate appearance using peak signal-to-noise ratio (PSNR), structural similarity index (SSIM), and perceptual similarity (LPIPS). For geometry comparison, we compute Chamfer Distance (CD), Point to Surface (P2S) distance, and normal consistency (NC). For details see Sup.~Mat. 


\noindent\textbf{Baselines.}
We conduct experiments on current state-of-the-art methods for single-view human reconstruction, including Human3Diffusion~\cite{xue2024human3diffusion}, SIFU \cite{zhang2024sifu}, SiTH \cite{ho2024sith}, and a concurrent method, PSHuman \cite{li2024pshuman}. More details about baselines can be found in Sup.~Mat.


\begin{figure*}[t]
    \centerline{    \includegraphics[width=\textwidth]{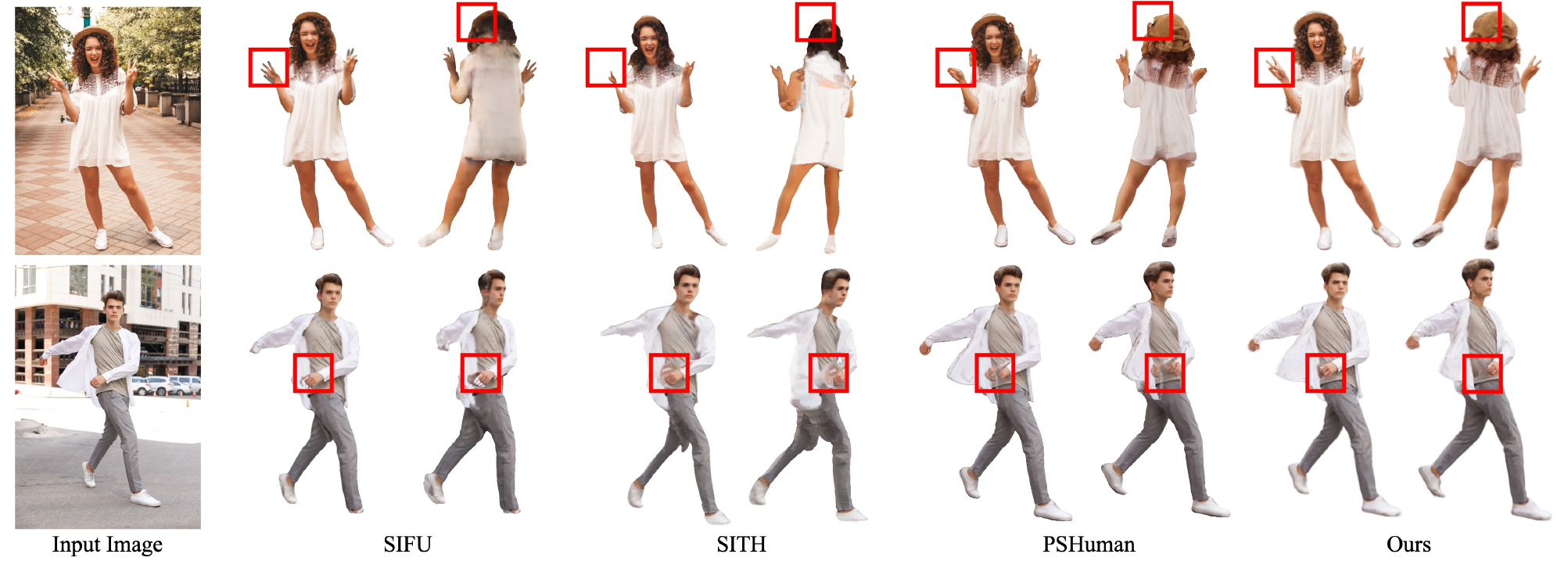}}
    \vspace{-0.1in}
    \caption{\textbf{Qualitative comparison to SotA methods on in-the-wild images:} Ours outperforms baselines on in-the-wild images by generating more plausible back/side views,  reconstructing finer details such as fingers and hats, and avoids artifacts due to self-occlusion.  }
    \label{fig:comparison_wild}
\end{figure*}

\subsection{Comparison to SotA.}

\tabref{tab:comparison} summarizes our quantitative comparisons.
Since Human3Diffusion~\cite{xue2024human3diffusion} only supports low-resolution rendering, we mainly compare it qualitatively in \figref{fig:comparison}.   \tabref{tab:comparison} shows that our method largely outperforms other baselines in appearance, especially in PSNR. As shown in \figref{fig:comparison}, our method generates overall sharper images with more details. Our method also reconstructs better geometry both quantitatively and qualitatively with better local geometric detail. 
Here, we discuss the main reason for the improvements:

\noindent\textbf{Side views.}
The improvements of appearance and geometry are particularly pronounced for side views. This is because most baselines rely on 2D multi-view diffusion models to hallucinate side views, which are not 3D consistent. 
In contrast, our learned generative avatar model provides additional appearance and geometry constraints on multi-view consistency, yielding better results.

\begin{figure}
    \centerline{    \includegraphics[width=1.0\linewidth]{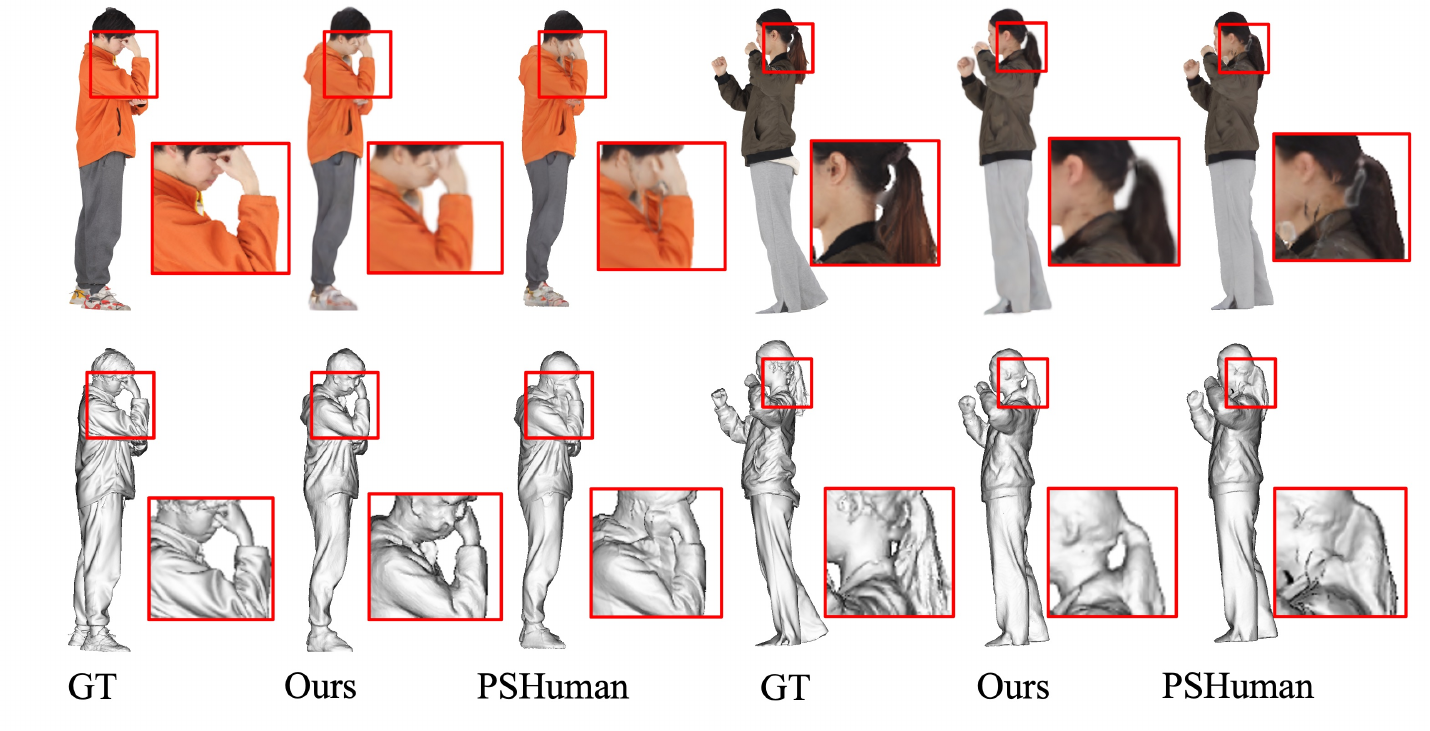}}
    \vspace{-0.1in}
    \caption{\textbf{Qualitative comparison to PSHuman on fine-sclae structures.} Our method reconstructs complex topologies, like a ponytail, that deviate from the body topology.}
    \label{fig:topology}
\end{figure}

\noindent\textbf{Self-occlusion.}
With a single input view, self-occlusion inevitably happens in the arm and hand regions due to the articulated nature of the human body as illustrated in \figref{fig:comparison}. In the first example, despite the use of the SMPL body prior, all previous methods fail to accurately reconstruct the left hand and arm. This is because these baselines tend to overfit to the input view, leading to incomplete reconstructions in occluded regions. In contrast, our method effectively reconstructs occluded arms and hands by leveraging the 3D appearance prior from our model. In the second example, the baselines exhibit artifacts on the clothing due to color misalignment from the arm, which is exacerbated by self-occlusion and depth ambiguity. Instead, our method reconstructs both the arm and clothing, despite the occlusion. 

\noindent\textbf{Topology changes.}
Figure \ref{fig:topology} compares our method with the concurrent PSHuman~\cite{li2024pshuman} on two more challenging subjects. Due to fixed topology of the template mesh, PSHuman fails to model areas between the arm and face, as well as the ponytail. These artifacts are present in other SMPL-based methods~\cite{zhang2024sifu}. In contrast, our method reconstructs 3D Gaussians and this flexible representation allows it to reconstruct more complex structures.


\begin{figure}
    \centerline{\includegraphics[width=1.0\linewidth]{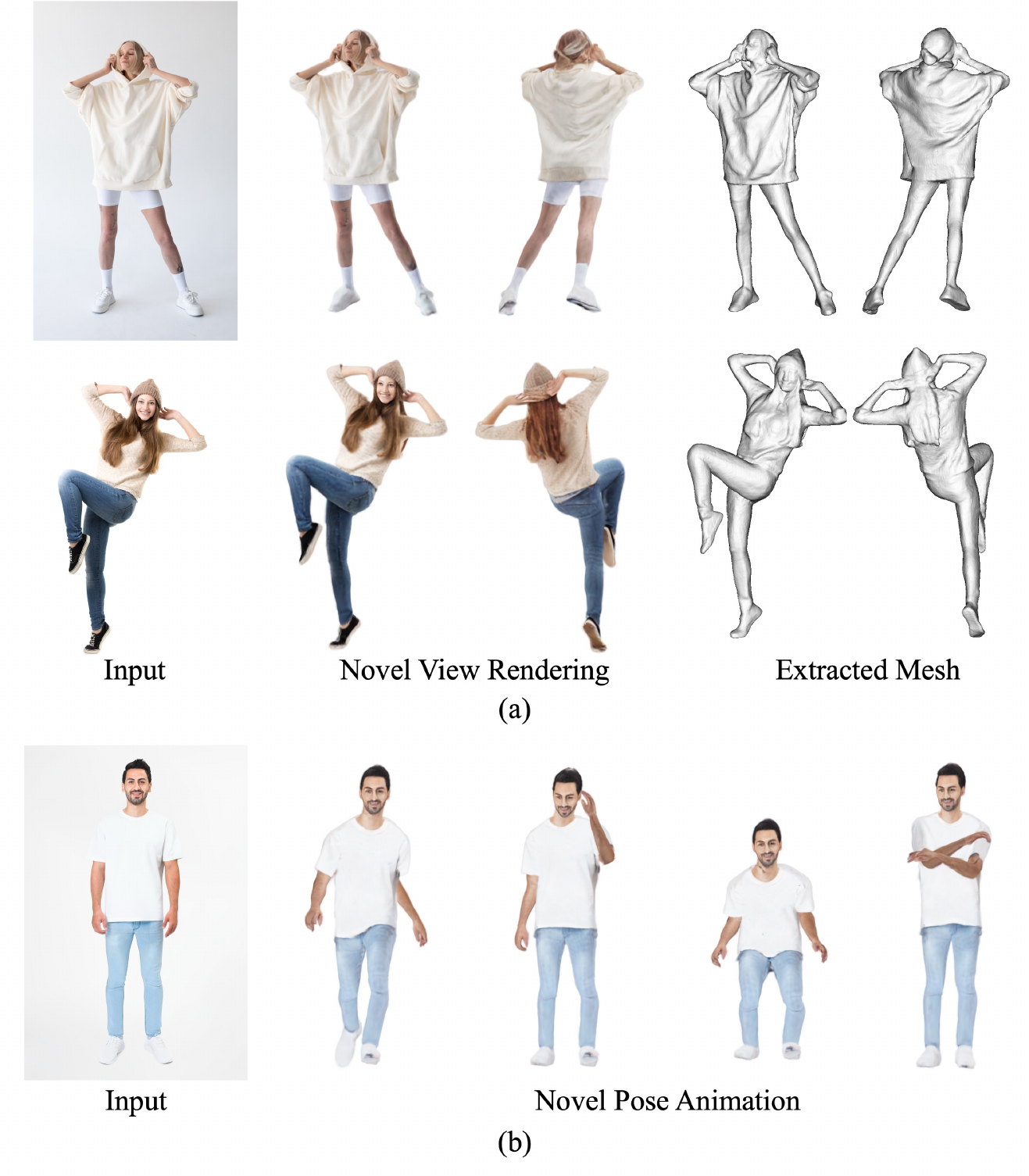}}
    \vspace{-0.1in}
    \caption{\textbf{Qualitative results on in-the-wild images}. (a)  From an in-the-wild image with challenging poses and clothing, our method reconstructs a high-quality Gaussian avatar, that enables realistic novel view synthesis and detailed geometry reconstruction.   (b) The resulting avatar can be animated with SMPL-X poses. }
    \label{fig:in_the_wild}
\end{figure}

\begin{table*}[t] 
    \centering
    \small
    \renewcommand{\arraystretch}{1.2} 
    \setlength{\tabcolsep}{2.5pt} 
  
    \begin{tabular}{l c c c c c c | c c c c c c}
        \toprule
        Method & \multicolumn{6}{c|}{THuman2.1} & \multicolumn{6}{c}{CustomHuman} \\
        \cmidrule(lr){2-7} \cmidrule(lr){8-13}
        & PSNR↑ & SSIM↑ & LPIPS↓ & CD↓ & P2S↓ & NC↑ & PSNR↑ & SSIM↑ & LPIPS↓ & CD↓ & P2S↓ & NC↑ \\ 
        \midrule
        SIFU\cite{zhang2024sifu} & 17.5271 & 0.9219 & 0.1019 & 2.6220& 2.4450 & 0.787 & 16.0198& 0.9056& 0.1146& 2.3717 & 2.2206 & 0.809\\
        SiTH \cite{ho2024sith}& 19.4020 & 0.9344 & 0.0796 & 2.2370 & 1.8459 & 0.808 & 17.7986 & 0.9211 & 0.0921 & 2.9142 & 2.0426 & 0.788  \\
        PSHuman \cite{li2024pshuman} & 19.9595 & 0.9350 & 0.0778 & 1.4128 & 1.2320 & 0.837 & 18.6704 & 0.9223 & 0.0850 & 1.9197 & \textbf{1.4695} & 0.828 \\
        Ours & \textbf{24.0926} & \textbf{0.9455} & \textbf{0.0732} & \textbf{1.3608} & \textbf{1.2226} & \textbf{0.850} & \textbf{23.4383} & \textbf{0.9351} & \textbf{0.0791} & \textbf{1.8086} & 1.4821 & \textbf{0.834} \\
        \bottomrule
    \end{tabular}
\vspace{-0.05in}
      \caption{ \textbf{Quantitative comparison with SotA Methods on Thuman2.1 and CustomHuman.} Our method outperforms other baselines by a large margin on appearance and also demonstrates a clear improvement in geometry.}
    \label{tab:comparison}
\end{table*}

\subsection{In-the-wild Performance}

Figure \ref{fig:in_the_wild}(a) shows qualitative results on in-the-wild images.
\model generalizes to loose clothing and challenging poses.
The reconstructed Gaussian avatar can be posed or animated (\figref{fig:in_the_wild}(b)), because it is based on SMPL-X. 
Figure \ref{fig:comparison_wild} compares results of SotA methods on in-the-wild images.
Both SiTH and SIFU struggle to produce reasonable reconstructions, often generating blurry back views and unrealistic appearance. PSHuman works better, but fails to capture fine details such as fingers and hats, while also introducing artifacts on clothing. In contrast, by leveraging the learned generative avatar prior, our \model model faithfully reconstructs Gaussian avatars in this challenging setting.



\subsection{Ablation Study}
Since the generative avatar prior is the key to our method, we focus on ablations that evaluate its effectiveness. 
An analysis of pose optimization,  unconditional generation, and robustness to the number of views appears in Sup.~Mat. All experiments are conducted on the CustomHuman Dataset. 

\begin{table}[h]
    \centering
    \small
    \renewcommand{\arraystretch}{1.2} 
    \setlength{\tabcolsep}{8pt} 
    
    \begin{tabular}{l c c c}
        \toprule
        Method & PSNR ↑ & SSIM ↑ & LPIPS ↓ \\
        \midrule
        w/o Initialization & 22.6996 & 0.9278 & 0.0891 \\
        w/o Avatar Prior & 22.5838 & 0.9288 & 0.0863\\
        Full Model & \textbf{23.4383} & \textbf{0.9351} & \textbf{0.0791} \\
        \bottomrule
    \end{tabular}
    \vspace{-0.05in}
    \caption{\textbf{Ablation.} We compare our method with ablated
baselines in which we remove the generative avatar prior for initialization and regularization. }
    \label{tab:ablation}
    
\end{table}

\begin{figure}
    \centerline{    \includegraphics[width=1.0\linewidth]{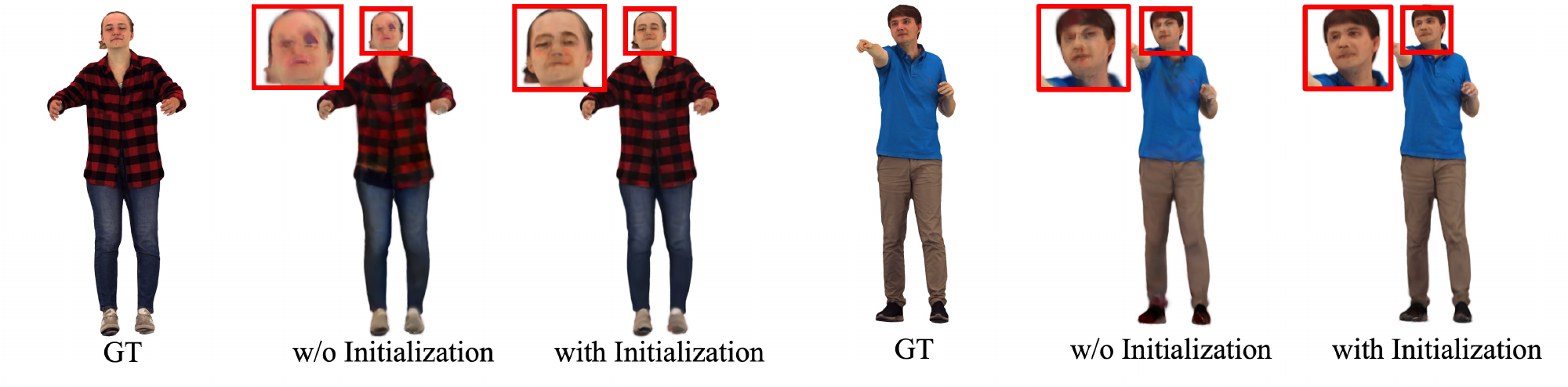}}
    \vspace{-0.1in}
    \caption{\textbf{Ablation of Initialization.} 
    The good initialization provided by our generative avatar model enhances appearance quality, reducing blurriness and producing a more detailed face}
    \label{fig:initialization}
\end{figure}

\noindent\textbf{Effect of initialization.}
Our generative avatar plays a crucial role in initializing the fitting process. We compare our method against ablated versions where the latent code is initialized randomly. 
As shown in \tabref{tab:ablation},  \model improves all appearance metrics. 
Furthermore, qualitative results in \figref{fig:initialization} demonstrate that, without proper initialization, the reconstructed appearance becomes blurry, particularly in the face region. This degradation occurs because the optimization process converges to a poor local minimum. In contrast, our method produces sharper and more accurate facial reconstructions, highlighting the importance of a well-initialized latent code.

\begin{figure}
    \centering
    \includegraphics[width=1.0\linewidth]{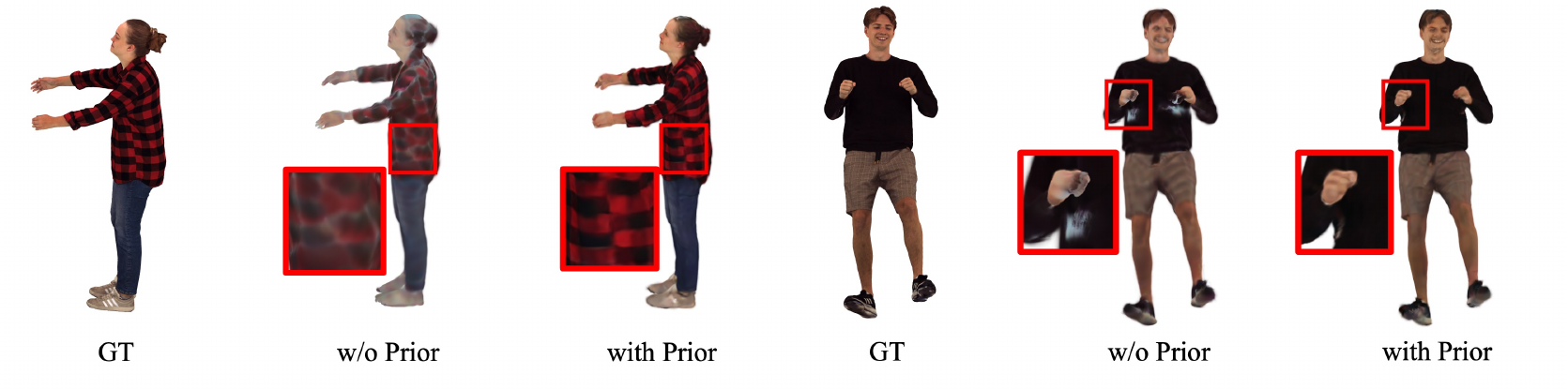}
    \caption{\textbf{Ablation of generative avatar prior.} The generative avatar model serves as an important 3D regularization to ensure 3D consistency and inpaint missing regions.}
    \label{fig:effect_prior}
\end{figure}

\noindent\textbf{Effect of generative avatar prior.} To evaluate the effect of the generative avatar prior, we create an ablated version of our method that directly optimizes SMPL-anchored Gaussians without using the learned decoder and latent diffusion model.  
\tabref{tab:ablation} shows that the generative avatar prior is important.
In the first example of \figref{fig:effect_prior}, removing the generative avatar prior results in a blurry side view and introduces a visible white crack between the front and back. This artifact arises due to the 3D inconsistency of images generated by the multi-view diffusion model. In contrast, our method enforces better 3D consistency. The second example in \figref{fig:effect_prior} highlights another issue without our model.  The ablated baseline tends to produce artifacts in occluded regions. In comparison, incorporating the generative prior enables our model to inpaint the missing areas effectively, resulting in a more natural and complete appearance. 
\section{Conclusion}

In this paper, we propose \model, a novel approach for reconstructing Gaussian avatars from a monocular image. 
Unlike previous methods that rely solely on multi-view diffusion, we integrate a 3D generative avatar model as a complementary prior, ensuring 3D consistency by projecting images into its latent space and enforcing both 3D appearance and geometry constraints.  We formulate Gaussian avatar creation as model inversion by fitting the generative avatar model to synthetic images from 2D diffusion models. 
Our method sets a new state-of-the-art in reconstruction quality and 3D consistency, generalizing well to in-the-wild images while producing animatable avatars without post-processing. Limitations and discussions appear in Sup.~Mat.

\vspace{-0.2em}
\boldparagraph{Acknowledgements} Andreas Geiger was supported by the ERC Starting Grant LEGO-3D (850533) and the DFG EXC number 2064/1 - project number 390727645. We thank Chengwei Zheng, Chen Guo, Juan Zarate, Zhiyi Chen and Jiarui Gao for their help and feedback. 

\noindent\textbf{Disclosure.}
While MJB is a co-founder and Chief Scientist at Meshcapade, his research in this project was performed solely at, and funded solely by, the Max Planck Society.

{\small
\bibliographystyle{ieeenat_fullname}
\bibliography{main}
}

\end{document}